\documentclass[10pt,twocolumn,letterpaper]{article}
\pdfoutput=1
\usepackage{cvpr}
\usepackage{times}
\usepackage{epsfig}
\usepackage{graphicx}
\usepackage{amsmath}
\usepackage{amssymb}
\usepackage{array}
\usepackage{enumitem}
\usepackage{multirow}
\usepackage[footnotesize]{subfigure}

\usepackage[lined,linesnumbered,ruled]{algorithm2e}

\SetCommentSty{mycommfont}
\DeclareMathOperator*{\argmax}{arg\,max}

\DeclareMathOperator*{\softmax}{softmax}

\usepackage[pagebackref=true,breaklinks=true,letterpaper=true,colorlinks,bookmarks=false]{hyperref}

\cvprfinalcopy 

\def\cvprPaperID{2337} 

\cvprfinalcopy 
\begin{document}

\title{Grounding Referring Expressions in Images by Variational Context}

\author{Hanwang Zhang$^{1}$\thanks{Hanwang was a research scientist at Columbia University.}  ~~~~~~~~Yulei Niu$^{2}$\thanks{Yulei was a visiting student at Columbia University.}   ~~~~~~~~Shih-Fu Chang$^3$\\
	$^1$Nanyang Technological University, $^2$Renmin University of China, $^3$Columbia University, \\
	hanwangzhang@ntu.edu.sg; niu@ruc.edu.cn; shih.fu.chang@columbia.edu\\
}

\maketitle

\begin{abstract}
We focus on grounding (i.e., localizing or linking) referring expressions in images, e.g., ``largest elephant standing behind baby elephant''.  This is a general yet challenging vision-language task since it does not only require the localization of objects, but also the multimodal comprehension of context --- visual attributes (e.g., ``largest'', ``baby'') and relationships (e.g., ``behind'') that help to distinguish the referent from other objects, especially those of the same category. Due to the exponential complexity involved in modeling the context associated with multiple image regions, existing work oversimplifies this task to pairwise region modeling by multiple instance learning. In this paper, we propose a variational Bayesian method, called Variational Context, to solve the problem of complex context modeling in referring expression grounding. Our model exploits the reciprocal relation between the referent and context, i.e., either of them influences estimation of the posterior distribution of the other, and thereby the search space of context can be greatly reduced. We also extend the model to unsupervised setting where no annotation for the referent is available. Extensive experiments on various benchmarks show consistent improvement over state-of-the-art methods in both supervised and unsupervised settings. The code is available at \url{https://github.com/yuleiniu/vc/}.
\end{abstract}
\vspace{-4mm}
\section{Introduction}
\begin{figure}
	\centering
	\includegraphics[width=.85\linewidth]{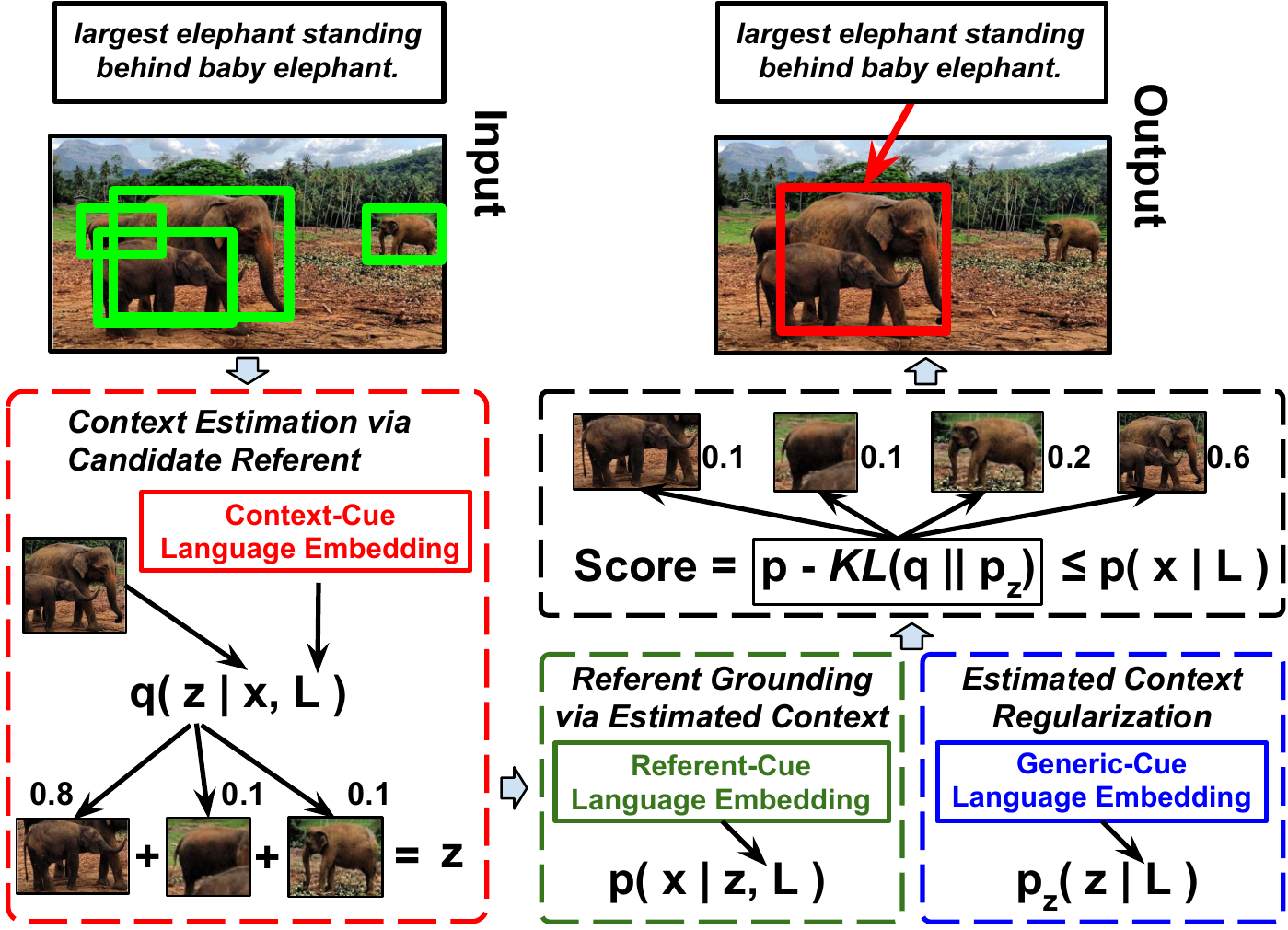}
	\caption{The proposed Variational Context model. Given an input referring expression and an image with region proposals, we localize the referent as output. We develop a grounding score function, with the variational lower-bound composed by three cue-specific multimodal modules, indicated by the description in the dashed color boxes.}
\vspace{-5mm}
\label{fig:1}
\end{figure}

Grounding natural language in visual data is a hallmark of AI, since it establishes a communication channel between humans, machines, and the physical world, underpinning a variety of multimodal AI tasks such as robotic navigation~\cite{thomason2017guiding}, visual Q\&A~\cite{antol2015vqa,li2017visual,zhao2017video}, and visual chatbot~\cite{visdial_rl}. Thanks to the rapid development in deep learning-based CV and NLP, we have witnessed promising results not only in grounding nouns (\eg, object detection~\cite{redmon2016yolo9000}), but also short phrases (\eg, noun phrases~\cite{plummer2016phrase} and relations~\cite{zhang2016vtranse,sun2017domain}). However, the more general task: grounding referring expressions~\cite{mao2016generation}, is still far from resolved due to the challenges in understanding of both language and scene compositions~\cite{golland2010game}. As illustrated in Figure~\ref{fig:1}, given an input referring expression ``largest elephant standing behind baby elephant'' and an image with region proposals, a model that can only localize ``elephant'' is not satisfactory as there are multiple elephants. Therefore, the key for referring expression grounding is to comprehend and model the context. Here, we refer to \emph{context} as the visual objects (\eg, ``elephant''), attributes (\eg, ``largest'' and ``baby''), and relationships (\eg, ``behind'') mentioned in the expression that help to distinguish the referent from other objects.

One straightforward way of modeling the relations between the referent and context is to: 1) use external syntactic parsers to parse the expression into entities, modifiers, and relations~\cite{schuster2015generating}, and then 2) apply visual relation detectors to localize them~\cite{zhang2016vtranse}. However, this two-stage approach is not practical due to the limited generalization ability of the detectors applied in the highly unrestricted language and scene compositions. To this end, recent approaches use multimodal embedding networks that jointly comprehend language and model the visual relations~\cite{nagaraja2016modeling,hu2016modeling}. Due to the prohibitively high cost of annotating both referent and context of referring expressions in images, multiple instance learning (MIL)~\cite{dietterich1997solving} is usually adopted in them to handle the weak supervision of the unannotated context objects, by maximizing the joint likelihood of every region pair. However, for a referent, the MIL framework essentially oversimplifies the number of context configurations of $N$ regions from $\mathcal{O}(2^N)$ to $\mathcal{O}(N)$. For example, to localize the ``elephant'' in Figure~\ref{fig:1}, we may need to consider the other three elephants all together as a multinomial subset for modeling the context such as ``largest'', ``behind'' and ``baby elephant''. 

In this paper, we propose a novel model called \emph{Variational Context} for grounding referring expressions in images. Compared to the previous MIL-based approaches~\cite{nagaraja2016modeling,hu2016modeling}, our model approximates the combinatorial context configurations with weak supervision using a variational Bayesian framework~\cite{kingma2013auto}. Intuitively, it exploits the reciprocity between referent and context, given either of which can help to localize the other. As shown in Figure~\ref{fig:1}, for each region $x$, we first estimate a coarse context $z$, which will help to refine the true localizations of the referent. This reciprocity is formulated into the variational lower-bound of the grounding likelihood $p(x|L)$, where $L$ is the text expression and the context is considered as a hidden variable $z$ (cf. Section~\ref{sec:3}). Specifically, the model consists of three multimodal modules: context posterior $q(z|x,L)$, referent posterior $p(x|z, L)$, and context prior $p_z(z|L)$, each of which performs a grounding task (cf. Section~\ref{sec:4_3}) that aligns image regions with a cue-specific language feature; each cue dynamically encodes different subsets of words in the expression $L$ that help the corresponding localization (cf. Section~\ref{sec:4_2}). 

Thanks to the reciprocity between referent and context, our model can not only be used in the conventional supervised setting, where there is annotation for referent , but also in the challenging unsupervised setting, where there is no instance-level annotation (\eg, bounding boxes) of both referent and context. We perform extensive experiments on four benchmark referring expression datasets: RefCLEF~\cite{kazemzadeh2014referitgame}, RefCOCO~\cite{yu2016modeling}, RefCOCO+~\cite{yu2016modeling}, and RefCOCOg~\cite{mao2016generation}. Our model consistently outperforms previous methods in both supervised and unsupervised settings. We also qualitatively show that our model can ground the context in the expression to the corresponding image regions (cf. Section~\ref{sec:5}).

\section{Related Work}\label{sec:2}
\textbf{Grounding Referring Expressions}.
Grounding referring expression is also known as referring expression comprehension, whose inverse task is called referring expression generation~\cite{mao2016generation}. Different from grounding phrases~\cite{plummer2015flickr30k,plummer2016phrase} and descriptive sentences~\cite{hu2016natural,rohrbach2016grounding}, the key for grounding referring expression is to use the context (or pragmatics in linguistics~\cite{thomas2014meaning}) to distinguish the referent from other objects, usually of the same category~\cite{golland2010game}. However, most previous works resort to use holistic context such as the entire image~\cite{mao2016generation, hu2016natural,rohrbach2016grounding} or visual feature difference between regions~\cite{yu2016modeling,yu2016joint}. Our model is similar to the works on explicitly modeling the referent and context region pairs~\cite{hu2016modeling, nagaraja2016modeling}, however, due to the lack of context annotation, they reduce the grounding task into a multiple instance learning framework~\cite{dietterich1997solving}. As we will discuss later, this framework is not a proper approximation to the original task. There are also studies on visual relation detection that detect objects and their relationships~\cite{lu2016visual,Dai_2017_CVPR,zhang2016vtranse,li2017vip,zhang2017ppr}. However, they are limited to a fixed-vocabulary set of relation triplets and hence are difficult to be applied in natural language grounding. Our cue-specific language feature is similar to the language modular network~\cite{hu2016modeling} that learns to decompose a sentence into referent/context-related words, which are different from other approaches that treat the expression as a whole~\cite{mao2016generation,luo2017comprehension,yu2016joint,liureferring}.

\textbf{Variational Bayesian Model vs. Multiple Instance Learning}.
Our proposed variational context model is in a similar vein of the deep neural network based variational autoencoder (VAE)~\cite{kingma2013auto}, which uses neural networks to approximate the posterior distribution of the hidden value $q(z|x)$, \ie, encoder, and the conditional distribution of the observation $p(x|z)$, \ie, decoder. VAE shows efficient and effective end-to-end optimization for the \emph{intractable} log-sum likelihood $\log\sum_z p(x,z)$ that is widely used in generative processes such as image synthesis~\cite{yan2016attribute2image} and video frame prediction~\cite{xue2016visual}. Considering the unannotated context as the hidden variable $z$, the referring expression grounding task can also be formulated into the above log-sum marginalization (cf. Eq.~\eqref{eq:2}). The MIL framework~\cite{dietterich1997solving} is essentially a sum-log approximation of the log-sum, \ie, $\sum_z\log p(x,z)$. To see this, the max-pooling function $\log\max_z p(x,z)$ used in~\cite{hu2016modeling} can be viewed as the sum-log $\sum_z\log p(x|z)p(z)$, where $p(z) = 1$ if $z$ is the correct context and 0 otherwise, indicating there is only one positive instance; maximizing the noisy-or function $\log(1-\prod_z(1-p(x,z)))$ used in~\cite{nagaraja2016modeling} is equivalent to maximize $\sum_{z}\log p(x,z)$, assuming there is at least one positive instance. However, due to the numerical property of the log function, this sum-log approximation will unnecessarily force every $(x,z)$ pair to explain the data~\cite{fox2012tutorial}. Instead, we use the variational Bayesian upper-bound to obtain a better sum-log approximation. Note that visual attention models~\cite{ba2014multiple,xu2015show} simplify the variational lower bound by assuming $p(z) = q(z|x)$; however, we explicitly use the KL divergence $KL(q(z|x)||p(z))$ in the lower bound to regularize the approximate posterior $q(z|x)$ not being too far from the prior $p(z)$.

\section{Variational Context}\label{sec:3}
In this section, we derive the variational Bayesian formulation of the proposed variational context model and the objective function for training and test.
\subsection{Problem Formulation}
The task of grounding a referring expression $L$ in an image $I$, represented by a set of regions $x\in\mathcal{X}$, can be viewed as a region retrieval task with the natural language query $L$. Formally, we maximize the log-likelihood of the conditional distribution to localize the referent region $x^*\in\mathcal{X}$:
\begin{equation}\label{eq:1}
x^* = \argmax_{x\in\mathcal{X}}\log p(x|L),
\end{equation}
where we omit the image $I$ in $p(x|I, L)$. 

As there is usually no annotation for the context, we consider it as a hidden variable $z$. Therefore, Eq.~\eqref{eq:1} can be rewritten as the following maximization of the log-likelihood of the conditional marginal distribution:
\begin{equation}\label{eq:2}
x^* = \argmax_{x\in\mathcal{X}}\log \sum\limits_{z}p(x,z|L).
\end{equation}
Note that $z$ is NOT necessary to be one region as assumed in recent MIL approaches~\cite{hu2016modeling,nagaraja2016modeling}, \ie, $z\in\mathcal{X}$. For example, the contextual objects ``surrounding elephants'' in ``a bigger elephant than the surrounding elephants'' should be composed by a multinomial subset of $\mathcal{X}$, resulting in an extremely large sample space that requires $\mathcal{O}(2^{|\mathcal{X}|})$ search complexity. Therefore, the marginalization in Eq~\eqref{eq:2} is intractable in general. 

To this end, we use the variational lower-bound~\cite{kingma2013auto} to approximate the marginal distribution in Eq.~\eqref{eq:2} as:
\begin{equation}\label{eq:3}
\begin{split}
&\log p(x|L) = \log \sum\limits_z p(x,z|L) \geq \mathcal{Q}(x,L)=\\
&\underbrace{\mathbb{E}_{z\sim q_{\phi}(z|x,L)}\log p_{\theta}(x|z,L)}_{\text{Localization }}-\underbrace{KL\left(q_\phi(z|x,L)||p_\omega(z|L)\right)}_{\text{Regularization}},
\end{split}
\end{equation}
where $KL(\cdot)$ is the Kullback-Leibler divergence, $\phi$, $\theta$, and $\omega$ are independent parameter sets for the respective distributions. As shown in Figure~\ref{fig:1}, the lower bound $\mathcal{Q}(x, L)$ offers a new perspective for exploiting the reciprocal nature of referent and context in referring expression grounding:
\\
\textbf{Localization}. This term calculates the localization score for $x$ given an estimated context $z$, using the referent-cue of $L$ parameterized by $\theta$. In particular, we design a new posterior $q_\phi(z|x,L)$ that approximates the true context prior $p(z|x, L)$, which models the context $z$ using the context-cue of $L$ parameterized by $\phi$. In the view of variational auto-encoder~\cite{kingma2013auto,sohn2015learning}, this term works in an encoding-decoding fashion: $q_\phi$ is the encoder from $x$ to $z$, and $p_\theta$ is the decoder from $z$ to $x$. 
\\
\textbf{Regularization}.  As $KL$ is non-negative, maximizing $\mathcal{Q}(x,L)$ would encourage that the posterior $q_\phi$ is similar to the prior $p_\omega$, \ie, the estimated context $z$ sampled from $q_\phi (z|x, L)$ should not be too far from the referring expression, which is modeled by $p_\omega(z|L)$ with the generic-cue of $L$ parameterized by $\omega$. This term is necessary as the estimated $z$ could be overfitted to region features that are inconsistent with the visual context described in the expression. 
\begin{figure*}
	\centering
	\includegraphics[width=1\linewidth]{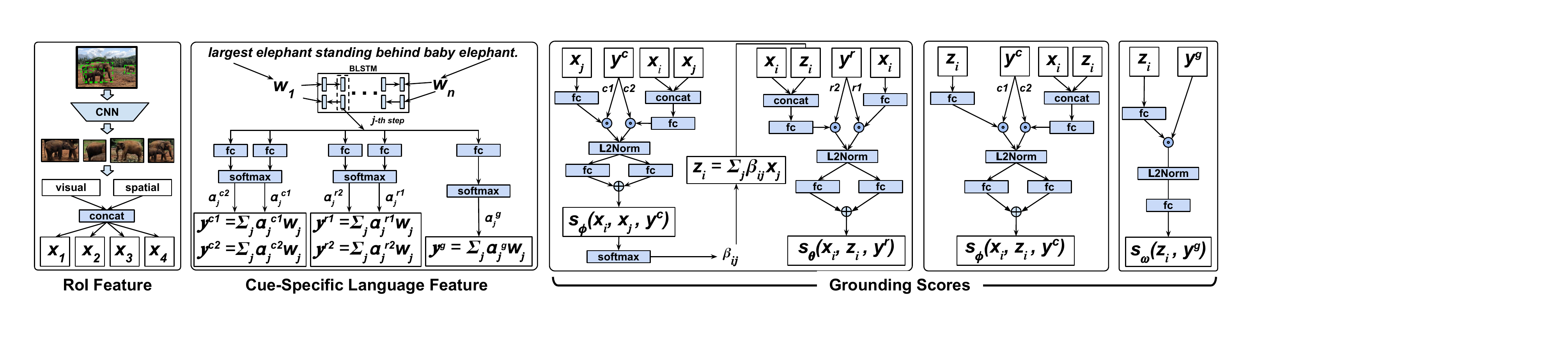}
	\caption{The architecture of the proposed Variational Context model. It consists of a region feature extraction module (Section~\ref{sec:4_1}, and a language feature extraction module (Section~\ref{sec:4_2}), and three grounding modules (Section~\ref{sec:4_3}). It can be trained in an end-to-end fashion with the input of a set of image regions and a referring expression, using the supervised loss ( Eq.~\eqref{eq:7}) or the unsupervised loss (Eq.~\eqref{eq:8}). fc: fully-connected layer. concat: vector concatenation. L2Norm: L2 normalization layer. $\odot$: element-wise vector multiplication. $\oplus$: add.}
\vspace{-4mm}
\label{fig:2}
\end{figure*}

\vspace{-2mm}
\subsection{Training and Test}
\textbf{Deterministic Context}. The lower-bound $\mathcal{Q}(x, L)$ transforms the intractable log-sum in Eq.~\eqref{eq:2} into the efficient sum-log in Eq.~\eqref{eq:3}, which can be optimized by using Monte Carlo unbiased gradient estimator such as REINFORCE~\cite{williams1992simple}. However, due to that $\phi$ is dependent on the sampling of $z$ over $\mathcal{O}(2^{|\mathcal{X}|})$ configurations, its gradient variance is large. To this end, we implement $q_\phi(z|x,L)$ as a differentiable but biased encoder:
\begin{equation}\label{eq:4} 
z = f(x,L) = \sum\limits_{x'\in\mathcal{X}}x'\cdot q_\phi(x'|x,L), 
\end{equation}
where we slightly abuse $q_\phi$ as a score function such that $\sum_{x'}q_\phi(x'|x,L) = 1$. Note that this deterministic context can be viewed as applying the ``re-parameterization'' trick as in Variational Auto-Encoder~\cite{kingma2013auto}: rewriting $z\sim q_\phi(z|x,L)$ to $z = f(x,L;\epsilon), \epsilon\sim p(\epsilon)$, where the stochasticity of the auxiliary random variable $\epsilon$ comes from training samples $x\in\mathcal{X}(\epsilon)$.  A clear example is Adversarial Autoencoder~\cite{aae} which shows that such stochasticity achieves similar test-likelihood compared to other distributions such as Gaussian.

\textbf{Objective Function}. Applying Eq.~\eqref{eq:4} to Eq.~\eqref{eq:3}, we can rewrite $Q(x,L)$ into a function of only one sample estimation, which is a common practice in SGD:
\begin{equation}\label{eq:5}
\begin{split}
\!\!\!\!\!\mathcal{Q}(x,\! L) &\!=\! \log p_\theta(x|z,\! L)\!-\!\log q_\phi(z|x,\! L)\!+\!\log p_\omega(z|L).
\end{split}
\end{equation}In supervised setting where the ground truth of the referent is known, to distinguish the referent from other objects, we need to train a model that outputs a high $p(x|L)$ (\ie, $\mathcal{Q}(x,L)$), while maintaining a low $p(x'|L)$ (\ie, $\mathcal{Q}(x',L)$), whenever $x'\neq x$. Therefore, we use the so-called Maximum Mutual Information loss as in~\cite{mao2016generation} $-\log\{\mathcal{Q}(x,L)/\sum_{x'}\mathcal{Q}(x',L)\}$, where we do not need to explicitly model the distributions with normalizations; we use the following score function:
\begin{equation}\label{eq:6}
\mathcal{Q}(x,L)\propto\mathcal{S}(x,L) = s_\theta(x, L)-s_\phi(x, L)+s_\omega(x, L),
\end{equation}
where $z$ is omitted as it is a function of $x$ in Eq.~\eqref{eq:4}. $s_\theta$, $s_\phi$, and $s_\omega$ are the score functions (\eg, $p_\theta\propto s_\theta $) for $p_\theta$, $q_\phi$, and $p_\omega$, respectively. These functions will be detailed in Section~\ref{sec:4_3}. In this way, maximizing Eq.~\eqref{eq:5} is equivalent to minimizing the following softmax loss:
\begin{equation}\label{eq:7}
\mathcal{L}_{s} = -\log\softmax \mathcal{S}(x_{gt},L),
\end{equation}
where the softmax is over $x\in\mathcal{X}$ and $x_{gt}$ is the ground truth referent region.

Note that the reciprocity between referent and context can be extended to unsupervised learning, where neither of the referent and context has annotation. In this setting, we adopt the \emph{image-level} max-pooled MIL loss functions for unsupervised referring expression grounding:
\begin{equation}\label{eq:8}
\mathcal{L}_{u} = -\max\limits_{x\in\mathcal{X}}\log\softmax\mathcal{S}(x,L),
\end{equation}
where the softmax is over $x\in\mathcal{X}$. Note that the max-pooled MIL function is reasonable since there is only one ground truth referent given an expression and image training pair. 

At test stage, in both supervised and unsupervised settings, we predict the referent region $x^*$ by selecting the region $x\in\mathcal{X}$ with the highest score:
\begin{equation}\label{eq:9}
x^* = \argmax\limits_{x\in\mathcal{X}}\mathcal{S}(x,L),
\end{equation}

\section{Model Architecture}
The overall architecture of the proposed variational context model is illustrated in Figure~\ref{fig:2}.  Thanks to the deterministic context in Eq.~\eqref{eq:4}, the five modules in our model can be integrated into an end-to-end differentiable fashion. Next, we will detail the implementation of each module.  
\subsection{RoI Features}\label{sec:4_1}
Given an image with a set of Region of Interests (RoIs) $\mathcal{X}$, obtained by any off-the-shelf proposal generator~\cite{zitnick2014edge} or object detectors~\cite{liu2016ssd}, this module extracts the feature vector $\mathbf{x}_i$ for every RoI. In particular, $\mathbf{x}_i$ is the concatenation of visual feature $\mathbf{v}_i$ and spatial feature $\mathbf{p}_i$. For $\mathbf{v}_i$, we can use the output of a pre-trained convolutional network (cf. Section~\ref{sec:5}). If the object category of each RoI is available, we can further utilize the comparison between the referent and other objects to capture the visual difference such as ``the largest/baby elephant''. Specifically, we append the visual difference feature~\cite{yu2016modeling} $\delta \mathbf{v}_i = \frac{1}{n}\sum_{j\neq i}\frac{\mathbf{v}_i-\mathbf{v}_j}{||\mathbf{v}_i-\mathbf{v}_j||}$ to the original $\mathbf{v}_i$ visual feature, where $n$ is the number of objects chosen for comparison (\eg, the number of RoI in the same object category). For spatial feature, we use the 5-d spatial attributes $\mathbf{p}_i = [\frac{x_{tl}}{W}, \frac{y_{tl}}{H}, \frac{x_{br}}{W}, \frac{y_{br}}{H}, \frac{w\cdot h}{W\cdot H}]$, where $x$ and $y$ are the coordinates the top left (tl) and bottom right (br) RoI of the size $w\times h$, and the image is of the size $W\times H$.

\subsection{Cue-Specific Language Features}\label{sec:4_2}
The cue-specific language feature representation for a referring expression is inspired by the attention weighted sum of word vectors~\cite{hu2016modeling, lu2016hierarchical,bahdanau2014neural}, where the weights are parameterized by context-cue $\phi$, referent-cue $\theta$, and generic-cue $\omega$. The context-cue language feature $\mathbf{y}^c = [\mathbf{y}^{c1}, \mathbf{y}^{c2}]$ is a concatenation of $\mathbf{y}^{c1}$ for language-vision association between \emph{single} RoI and the expression, and $\mathbf{y}^{c2}$ for the association between \emph{pairwise} RoIs; the referent-cue language feature $\mathbf{y}^r$ can be represented in a similar way to $\mathbf{y}^c$; the generic-cue language feature $\mathbf{y}^g$ is only for single RoI association as it is an independent prior. The weights of each cue are calculated from the hidden state vectors of a 2-layer bi-directional LSTM (BLSTM)~\cite{schuster1997bidirectional}, scanning through the expression. The hidden states encode forward and backward compositional semantic meanings of the sentences, beneficial for selecting words that are useful for single and pairwise associations. Specifically, suppose $\mathbf{h}_j$ as the 4,000-d concatenation of forward and backward hidden vectors of the $j$-th word, without loss of generality, the word attention weight $\alpha_j$ and the language feature $\mathbf{y}$ for single/pairwise association of any cue can be calculated as:
\begin{equation}\label{eq:10}
\mathbf{m}_j = \textrm{fc}(\mathbf{h}_j), \alpha_{j} = \textrm{softmax}_j(\mathbf{m}_{j}),\mathbf{y} = \sum\nolimits_{j}\alpha_j\mathbf{w}_j,
\end{equation}
where $\mathbf{w}_j$ is a 300-d vector. Note that the BLSTM module can be jointly trained with the entire model. 

\begin{figure}
	\centering
	\includegraphics[width=1\linewidth]{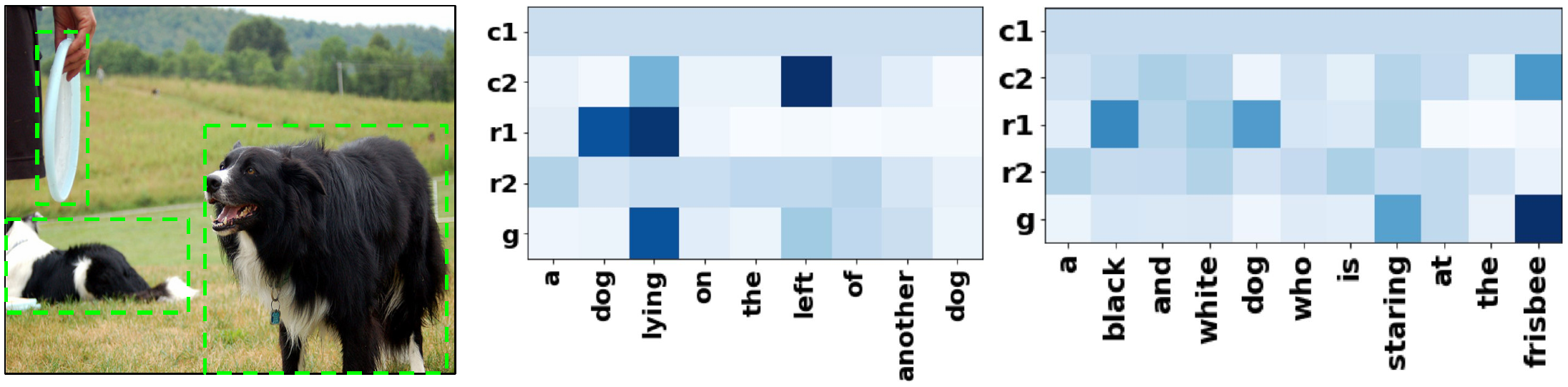}
	\caption{Two qualitative examples of the cue-specific language feature word weights. Darker color indicates higher weights. c/r+1/2: context/referent-cue + single/pairwise. }
\label{fig:3}
\vspace{-4mm}
\end{figure}

Figure~\ref{fig:3} shows that the cue-specific language features dynamically weight words in different expressions. We can have two interesting observations. First, c1 is almost uniform while c2 is highly skewed; although r2 is more skewed than c1, it is still less skewed than r1. This is reasonable since: 1) without ground-truth, individual score (c1) does not help much for context estimation from scratch; context is more easily found by the pairwise score (c2) induced by relationships or other objects (\eg, ``left'' or ``frisbee''); 2) in referent grounding with ground truth, individual score (r1) is sufficient (\eg, ``dog lying'' and ``black white dog'') and pairwise score (r2) is helpful; 3) g is adaptive to the number of object categories in the expression, \ie, if the context object is of the same category as the referent, g weighs descriptive or relationship words higher (\eg, ``lying, standing, left''), and nouns higher (\eg, ``frisbee''), otherwise; moreover, it demonstrates that the deterministic guess of $z$ in Eq.~\eqref{eq:4} is meaningful.

\subsection{Score Functions}\label{sec:4_3}
For any image and expression pair, given the RoI feature $\mathbf{x}_i$, and the cue-specific language feature $\mathbf{y}^c$, $\mathbf{y}^r$, and $\mathbf{y}^g$, we implement the final grounding score in Eq.~\eqref{eq:6} as:
\begin{equation}\label{eq:11x}
\begin{split}
\mathbf{z}_i = \sum\nolimits_j \textrm{softmax}_j&\left(s_\phi(\mathbf{x}_i,\mathbf{x}_j,\mathbf{y}^c)\right)\mathbf{x}_j,\\
s_\theta(x,L) &\leftarrow s_\theta(\mathbf{x}_i,\mathbf{z}_i,\mathbf{y}^r),\\
s_\phi(x,L) &\leftarrow s_\phi(\mathbf{x}_i,\mathbf{z}_i,\mathbf{y}^c),\\
s_\omega(x,L) &\leftarrow s_\omega(\mathbf{z}_i,\mathbf{y}^g),
\end{split}
\end{equation}
where the right-hand side functions are defined as below.

\textbf{Context Estimation Score}: $s_\phi(\mathbf{x}_i,\mathbf{x}_j,\mathbf{y}^c)$. It is a score function for modeling the context posterior $q_\phi(z|x,L)$, \ie, given an RoI $\mathbf{x}_i$ as the candidate referent, we calculate the likelihood of any RoI $\mathbf{x}_j$ to be the context. We can also use this function to estimate the final context posterior score $s_\phi(\mathbf{x}_i,\mathbf{z}_i,\mathbf{y}^c)$. Specifically, the context estimation score is a sum of the single and pairwise vision-language association scores: $\mathbf{x}_j$ and $\mathbf{y}^{c1}$, $[\mathbf{x}_i,\mathbf{x}_j]$ and $\mathbf{y}^{c2}$. Each associate score is an fc output from the input of a normalized feature:
\begin{equation}\label{eq:11}
\begin{split}
&\mathbf{m}^{1}_{j} = \mathbf{y}^{c1} \odot \textrm{fc}(\mathbf{x}_j),~\mathbf{m}^{2}_{j} = \mathbf{y}^{c2} \odot \textrm{fc}([\mathbf{x}_i,\mathbf{x}_j]),\\
&\widetilde{\mathbf{m}}^1_{j} = \textrm{L2Norm}(\mathbf{m}^{1}_{j}),~\widetilde{\mathbf{m}}^{2}_{j}= \textrm{L2Norm}(\mathbf{m}^{2}_{j}),\\
&s_\phi(\mathbf{x}_i,\mathbf{x}_j,\mathbf{y}^c) = \textrm{fc}(\widetilde{\mathbf{m}}^1_{j})+\textrm{fc}(\widetilde{\mathbf{m}}^{2}_{j}),
\end{split}
\end{equation}
where the element-wise multiplication $\odot$ is an effective way for multimodal features~\cite{ba2014multiple}. According to Eq.~\eqref{eq:4}, we can obtain the estimated context $z$ as $\mathbf{z}_i = \sum\nolimits_j\beta_{j}\mathbf{x}_j$, where $\beta_{j} = \textrm{softmax}_j(s_\phi(\mathbf{x}_i,\mathbf{x}_j,\mathbf{y}^c))$.

\textbf{Referent Grounding Score}: $s_\theta(\mathbf{x}_i,\mathbf{z}_i,\mathbf{y}^r)$. After obtaining the context feature $\mathbf{z}_i$, we can use this score function to calculate how likely a candidate RoI $\mathbf{x}_i$ is the referent given the context $\mathbf{z}_i$. This function is similar to Eq.~\eqref{eq:11}.

\textbf{Context Regularization Score}: $s_\omega(\mathbf{z}_i,\mathbf{y}^g)-s_\phi(\mathbf{x}_i,\mathbf{z}_i,\mathbf{y}^c)$. As discussed in Eq.~\eqref{eq:6}, this function scores how likely the estimated context feature $\mathbf{z}_i$ is consistent with the content mentioned in the expression. In particular, $s_\omega(\mathbf{z}_i,\mathbf{y}^g)$ is only dependent on single RoI:
\vspace{-1mm}
\begin{equation}\label{eq:12}
\mathbf{m}_{i}\!\! =\!\! \mathbf{y}^g_i \odot \textrm{fc}(\mathbf{z}_i),
\widetilde{\mathbf{m}}_{i} \!\!=\!\! \textrm{L2Norm}(\mathbf{m}_i),s_\omega(\mathbf{z}_i,\mathbf{y}^g_i)\!\!= \!\!\textrm{fc}(\mathbf{m}_i).
\end{equation}
\vspace{-5mm}
\section{Experiment}\label{sec:5}
\subsection{Datasets}
We used four popular benchmarks for the referring expression grounding task.

\textbf{RefCOCO}~\cite{yu2016modeling}. It has 142,210 referring expressions for 50,000 referents (\eg, object instances) in 19,994 images from MSCOCO~\cite{lin2014microsoft}. The expressions are collected in an interactive way~\cite{kazemzadeh2014referitgame}. The dataset is split into train, validation, Test A, and Test B, which has 120,624, 10,834, 5,657 and 5,095 expression-referent pairs, respectively. An image contains multiple people in Test A and multiple objects in Test B.

\textbf{RefCOCO+}~\cite{yu2016modeling}. It has 141,564 expressions for 49,856 referents in 19,992 images from MSCOCO. The difference from RefCOCO is that it only allows appearances but no locations to describe the referents. The split is 120,191, 10,758, 5,726 and 4,889 expression-referent pairs for train, validation, Test A, and Test B respectively. 

\textbf{RefCOCOg}~\cite{mao2016generation}. It has 95,010 referring expressions for 49,822 objects in 25,799 images from MSCOCO. Different from RefCOCO and RefCOCO+, this dataset not collected in an interactive way and contains longer sentences containing both appearance and location expressions. The split is 85,474 and 9,536 expression-referent pairs for training and validation. Note that there is no open test split for RefCOCOg, so we used the hyper-parameters cross-validated on RefCOCO and RefCOCO+.

\textbf{RefCLEF}~\cite{kazemzadeh2014referitgame}. It contains 20,000 images with annotated image regions. It has some ambiguous (e.g. “anywhere”) phrases and mistakenly annotated image regions that are not described in the expressions. For fair comparison, we used the split released by~\cite{hu2016natural,rohrbach2016grounding}, \ie, 58,838, 6,333 and 65,193 expression-referent pairs for training, validation and test, respectively.

\begin{table*}[t]
\centering
\caption{Supervised grounding performances (Acc\%) of comparing methods on RefCOCO, RefCOCO+, and RefCOCOg. Note that~\cite{yu2016joint} reports slightly higher accuracies using ensemble models of Listener and Speaker. For fair comparison, we only report their single models.}
\label{tab:1}
\scalebox{.8}{
\begin{tabular}{|c|c|c|c|c|c|c|c|c|c|c|}
\hline
\multicolumn{2}{|c|}{}             & \multicolumn{6}{c|}{State-of-The-Arts}     & \multicolumn{3}{c|}{Our Baselines}   \\ \hline
Dataset                   & Split  & MMI~\cite{mao2016generation} & NegBag~\cite{nagaraja2016modeling} & Attr~\cite{liureferring} & CMN~\cite{hu2016modeling} & Speaker~\cite{yu2016joint} & Listener~\cite{yu2016joint} & VC w/o reg & VC w/o $\alpha$ & VC \\ \hline
\multirow{2}{*}{RefCOCO}  & Test A & 71.72    & 75.6           &  78.85         &  75.94  & 78.95  & 78.45  &    75.59      &  74.03 & \textbf{78.98}     \\ \cline{2-11} 
                          & Test B & 71.09    &  78.0          &   78.07        &  79.57   & 80.22 & 80.10   &  79.69 &   78.27  & \textbf{82.39}        \\ \hline
\multirow{2}{*}{RefCOCO+} & Test A & 58.42    &   ---         &    61.47       &  59.29 &  \textbf{64.60} &   63.34    &   60.76  &  57.61   &  62.56      \\ \cline{2-11} 
                          & Test B & 51.23    &  ---          &   57.22        &  59.34 & 59.62  &   58.91    &      60.14  & 54.37  & \textbf{62.90}        \\ \hline
RefCOCOg                  & Val    &  62.14   &  68.4          &   69.83        &  69.30   & 72.63  & 72.25   & 71.05   & 65.13   & \textbf{73.98}        \\ \hline
\hline
\multirow{2}{*}{RefCOCO(det)}  & Test A & 64.90    & 58.6           &  72.08         &  71.03   & 72.95 & 72.95   &     70.78 &  70.73   & \textbf{73.33}        \\ \cline{2-11} 
                          & Test B & 54.51    &    56.4        &   57.29        &  65.77   &  63.43 & 62.98  &  65.10  &  64.63  &\textbf{67.44}       \\ \hline
\multirow{2}{*}{RefCOCO+(det)} & Test A &54.03     &   ---         &  57.97         &   54.32  & \textbf{60.43} & 59.61   &     56.82 &  53.33   &    58.40     \\ \cline{2-11} 
                          & Test B &42.81     &  ---          &  46.20         &  47.76   &  48.74  & 48.44  &   51.30      & 46.88    & \textbf{53.18}   \\ \hline
RefCOCOg(det)                  & Val    & 45.85    &39.5          &   52.35        &  57.47   &  59.51 & 58.32  &     60.95    &  55.72  & \textbf{62.30}    \\ \hline
\end{tabular}
}
\end{table*}

\begin{figure*}[t]
	\centering
	\includegraphics[width=1\linewidth]{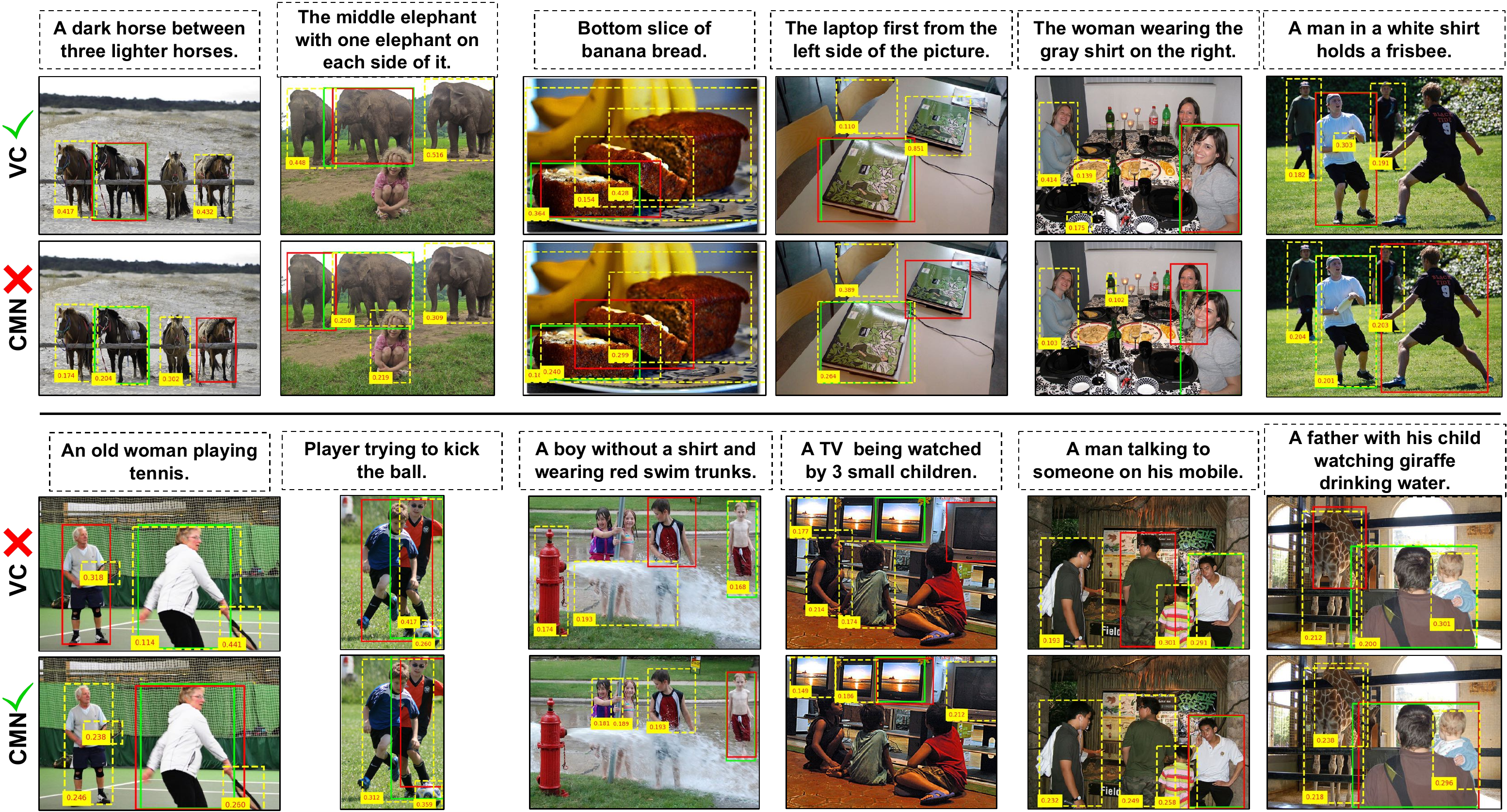}
	\caption{Qualitative results on RefCOCOg (det) showing comparisons between correct (green tick) and wrong referent grounds (red cross) by VC and CMN. The denotations of the bounding box colors are as follows. Solid red: referent ground; solid green: ground truth; dashed yellow: context ground. We only display top 3 context objects with the context ground probability $>0.1$. We can observe that VC has more reasonable context localizations than CMN, even in cases when the referent ground of VC fails.}
\vspace{-2mm}
\label{fig:4}
\end{figure*}

\subsection{Settings and Metrics}
We used an English vocabulary of 72,704 words contained in the GloVe pre-trained word vectors~\cite{pennington2014glove}, which was also used for the initialization of our word vectors. We used a ``unk'' symbol for the input word of the BLSTM if the word is out of the vocabulary; we set the sentence length to 20 and used ``pad'' symbol to pad expression sentence $<20$. For RoI visual features on RefCOCO, RefCOCO+, and RefCOCOg which have MSCOCO annotated regions with object categories, we used the concatenation of the 4,096-d fc7 output of a VGG-16 based Faster-RCNN network~\cite{ren2015faster} trained on MSCOCO and its corresponding 4,096-d visdiff feature~\cite{yu2016modeling}; although RefCLEF regions also have object categories, for fair comparison with~\cite{rohrbach2016grounding}, we did not use the visdiff feature. 

The model training is single-image based, with all referring expressions annotated. We applied SGD of 0.95-momentum with initial learning rate of 0.01, multiplied by 0.1 after every 120,000 iterations, up to 160,000 iterations. Parameters in BILSTM and fc-layers were initialized by Xavier~\cite{glorot2010understanding} with 0.0005 weight decay. Other settings were default in TensorFlow. Note that our model is trained without bells and whistles, therefore, other optimization tricks such as batch normalization~\cite{ioffe2015batch} and GRU~\cite{cho2014properties} are expected to further improve the results reported here. Besides the ground truth annotations, grounding to automatically detected objects is a more practical setting. Therefore, we also evaluated with the SSD-detected bounding boxes~\cite{liu2016ssd} on the four datasets provided by~\cite{yu2016joint}. A grounding is considered as correct if the intersection-over-union (IoU) of the top-1 scored region and the ground-truth object is larger than $0.5$. The grounding accuracy (a.k.a, P@1) is the fraction of correctly grounded test expressions.

\subsection{Evaluations of Supervised Grounding}
We compared our variational context model (VC) with state-of-the-art referring expression methods published in recent years, which can be categorized into: 1) generation-comprehension based such as MMI~\cite{mao2016generation}, Attr~\cite{liureferring}, Speaker~\cite{yu2016joint}, Listener~\cite{yu2016joint}, and SCRC~\cite{hu2016natural}; 2) localization based such as GroundR~\cite{rohrbach2016grounding}, NegBag~\cite{nagaraja2016modeling}, CMN~\cite{hu2016modeling}. Note that NegBag and CMN are MIL-based models. In particular, we used the author-released code to obtain the results of CMN on RefCLEF, RefCOCO, and RefCOCO+. 

From the results on RefCOCO, RefCOCO+, and RefCOCOg in Table~\ref{tab:1} and that on RefCLEF in Table~\ref{tab:2}, we can see that VC achieves the state-of-the-art performance. We believe that the improvement is attributed to the variational Bayesian modeling of context. First, on all datasets, except for the most recent reinforcement learning based~\cite{yu2016joint}, VC outperforms all the other sentence generation-comprehension methods that do not model context. Second, compared to VC without the regularization term in Eq.~\eqref{eq:3} (VC w/o reg), VC can boost the performance by around 2\% on all datasets. This demonstrates the effectiveness of the KL divergence for the prevention of the overfitted context estimation.  

In particular, we further demonstrate the superiority of VC over the most recent MIL-based method CMN. As illustrated in Figure~\ref{fig:4}, VC has better context comprehension in both of the language and image regions than CMN. For example, in the top two rows where VC is correct and CMN is wrong, for the grounding in the second column, CMN unnecessarily considers the ``girl'' as context but the expression only describes using ``elephant''; in the last column, CMN misses the key context ``frisbee''. Even in the failure cases where VC is wrong and CMN is correct, VC still localizes reasonable context. For example, in the fourth column, although CMN grounds the correct TV, but it is based on incorrect context of other TVs; while VC can predict the correct context ``children''. In addition, we observed that most of the cases that CMN is better than VC involves multiple humans. This demonstrates that VC is better at grounding objects of different categories. 

\begin{figure*}
	\centering
	\includegraphics[width=1\linewidth]{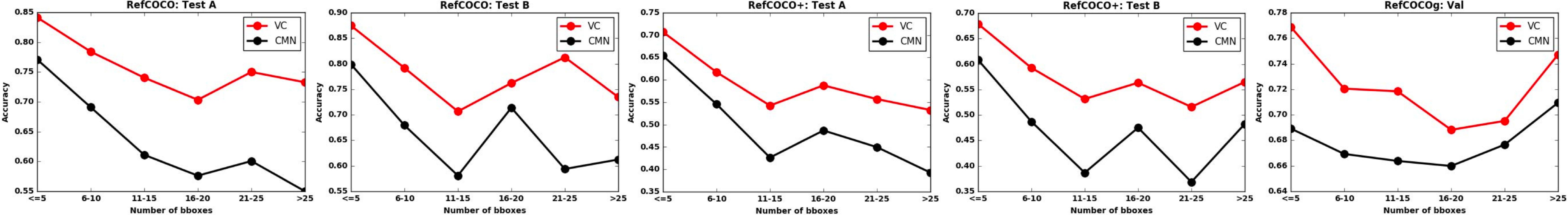}
	\caption{Performances of VC and CMN with different number of object bounding boxes on RefCOCO Test A \&B, RefCOCO+ Test A \& B, and RefCOCOg Val. Compared to CMN, we can see that VC is more effective in context modeling when the number of objects is large.}
\vspace{-4mm}
\label{fig:5}
\end{figure*}

VC is also effective in images with more objects. Figure~\ref{fig:5} shows the performances of VC and CMN with various number of bounding boxes. We can observe that VC considerably outperforms CMN over all bounding boxes numbers. Recall that context is the key to distinguish objects of the same category. In particular, on the Test A sets of RefCOCO and RefCOCO+ where the grounding is only about people, \ie, the same object category, the gap between VC and CMN is becoming larger as the box number increases. This demonstrates that MIL is ineffective in modeling context, especially when the number of image regions is large.

\subsection{Evaluations of Unsupervised Grounding}

\begin{table}
\centering
\caption{Performances (Acc\%) of supervised and unsupervised methods on RefCLEF. }
\label{tab:2}
\scalebox{.8}{
\begin{tabular}{|c|c|c|c|}
\hline
                       & Sup. & Sup. (det) & Unsup. (det) \\ \hline
SCRC~\cite{hu2016natural}    &  72.74    &     17.93       &   ---           \\ \hline
GroundR~\cite{rohrbach2016grounding} &  ---    &     26.93       &   10.70           \\ \hline
CMN~\cite{hu2016modeling} & 81.52 & 28.33 &---  \\ \hline\hline
VC                     &   \textbf{82.43}   &     \textbf{31.13}     &   14.11   \\ \hline
VC w/o $\alpha$      &  79.60    &    27.40        &   \textbf{14.50}           \\ \hline
\end{tabular}
}
\end{table}

\begin{table}[]
\centering
\caption{Unsupervised grounding performances (Acc\%) of comparing methods on RefCOCO, RefCOCO+, and RefCOCOg.}
\label{tab:3}
\scalebox{.8}{
\begin{tabular}{|c|c|c|c|c|}
\hline
Dataset                   & Split  & VC w/o reg & VC & VC w/o $\alpha$\\ \hline
\multirow{2}{*}{RefCOCO}  & Test A &  13.59  &  17.34  &  \textbf{33.29}  \\ \cline{2-5} 
                          & Test B &  21.65  &  20.98  &  \textbf{30.13}  \\ \hline
\multirow{2}{*}{RefCOCO+} & Test A &  18.79  &  23.24  &  \textbf{34.60}  \\ \cline{2-5} 
                          & Test B &  24.14  &  24.91  &  \textbf{31.58}  \\ \hline
RefCOCOg                  & Val    &  25.14  &  \textbf{33.79}  &  30.26     \\ \hline
\hline
\multirow{2}{*}{RefCOCO(det)}  & Test A & 17.14 & 20.91 & \textbf{32.68} \\ \cline{2-5} 
                               & Test B & 22.30 & 21.77 & \textbf{27.22}      \\ \hline
\multirow{2}{*}{RefCOCO+(det)} & Test A & 19.74 & 25.79 & \textbf{34.68}  \\ \cline{2-5} 
                               & Test B & 24.05 & 25.54 & \textbf{28.10}     \\ \hline
RefCOCOg(det)                  & Val    & 28.14 & \textbf{33.66} & 29.65     \\ \hline
\end{tabular}
}
\end{table}

\begin{figure}
	\centering
	\includegraphics[width=.8\linewidth]{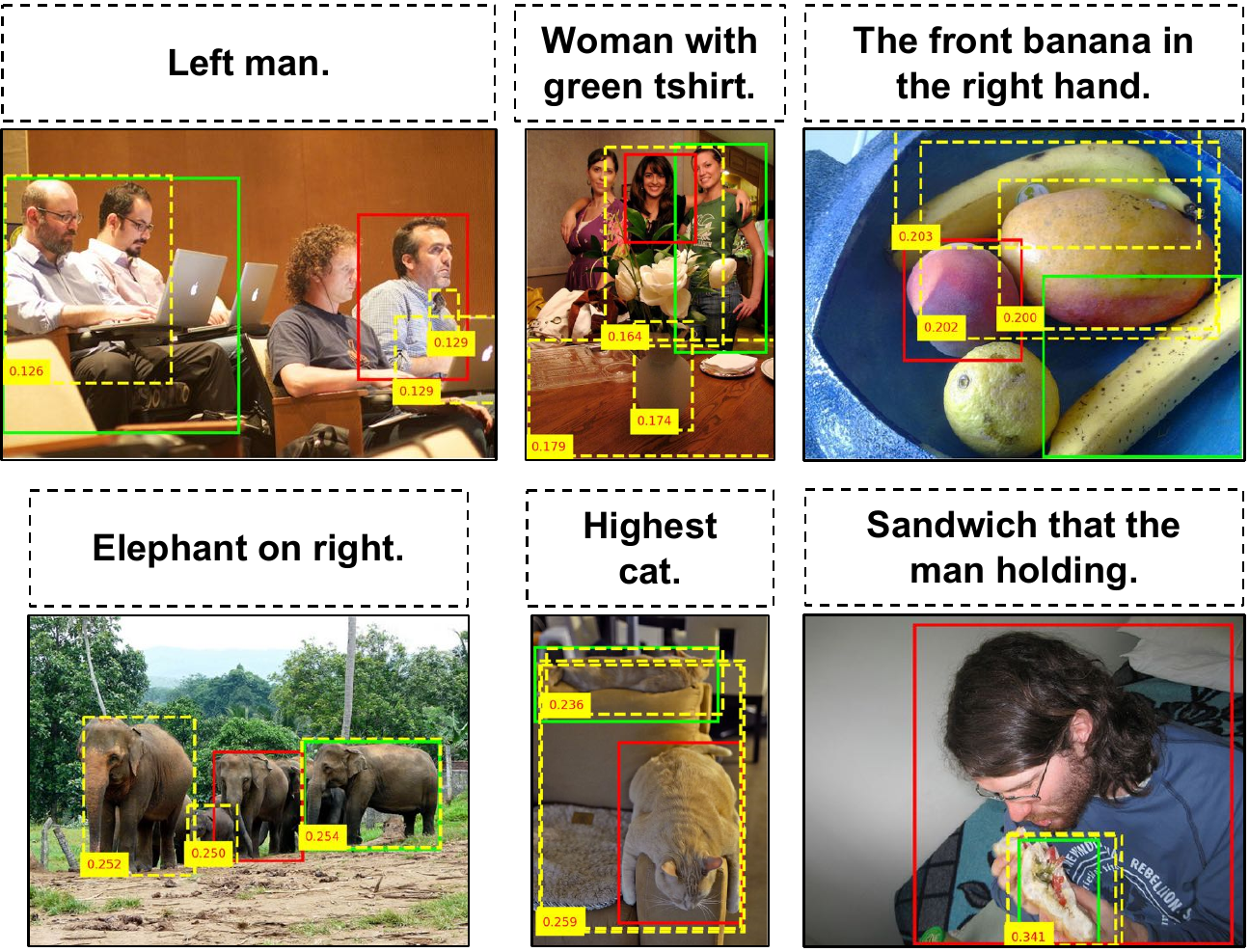}
	\caption{Common failure cases in unsupervised grounding with detected bounding boxes. From left to right: RefCOCO, RefCOCO+, and RefCOCOg. The failure is mainly to the challenging unsupervised relation modeling between referent and context.}
\vspace{-3mm}
\label{fig:7}
\end{figure}

We follow the unsupervised setting in GroundR~\cite{rohrbach2016grounding}. To our best knowledge, it is the only work on unsupervised referring expression grounding.  Note that it is also known as ``weakly supervised'' detection~\cite{zhang2017ppr} as there is still image-level ground truth (\ie, the referring expression). Table~\ref{tab:2} reports the unsupervised results on the RefCLEF. We can see that VC outperforms the state-of-the-art GroundR, which is a generation-comprehension based method. This demonstrates that using context also helps unsupervised grounding. As there is no published unsupervised results on RefCOCO, RefCOCO+, and RefCOCOg, we only compared our baselines on them in Table~\ref{tab:3}. We can have the following three key observations which highlight the challenges of unsupervised grounding:

\textbf{Context Prior}. VC w/o reg is the baseline without the KL divergence as a context regularization in Eq.~\eqref{eq:3}. We can see that in most of the cases, VC considerably outperforms VC w/o reg by over 2\%, even over 5\% on RefCOCO+ (det) and RefCOCOg (det). Note that this improvement is significantly higher than that in supervised setting (\eg, $<3\%$ as reported in Table~\ref{tab:1}). The reason is that the context estimation in Eq.~\eqref{eq:4} would be easier to be stuck in image regions that are irrelevant to the expression in unsupervised setting, therefore, context prior is necessary. 

\textbf{Language Feature}. Except on RefCOCOg, we consistently observed the \emph{ineffectiveness} of the cue-specific language feature in unsupervised setting, \ie, VC w/o $\alpha$ outperforms VC in Table~\ref{tab:2} and~\ref{tab:3}. Here $\alpha$ represents the cue-specific word attention. This is contrary to the observation in the supervised setting as listed in Table~\ref{tab:1}, where VC w/o $\alpha$ is consistently lower than VC. Note that without the cue-specific word attention $\alpha$ in Eq.~\eqref{eq:10}, the language feature is merely the average value of the word embedding vectors in the expression. In this way, VC w/o $\alpha$ does not encode any structural language composition as illustrated in Figure~\ref{fig:3}, thus, it is better for short expressions. However, when the expression is long in RefCOCOg, discarding the language structure still degrades the performance on RefCOCOg. 

\begin{figure}
	\centering
	\includegraphics[width=.9\linewidth]{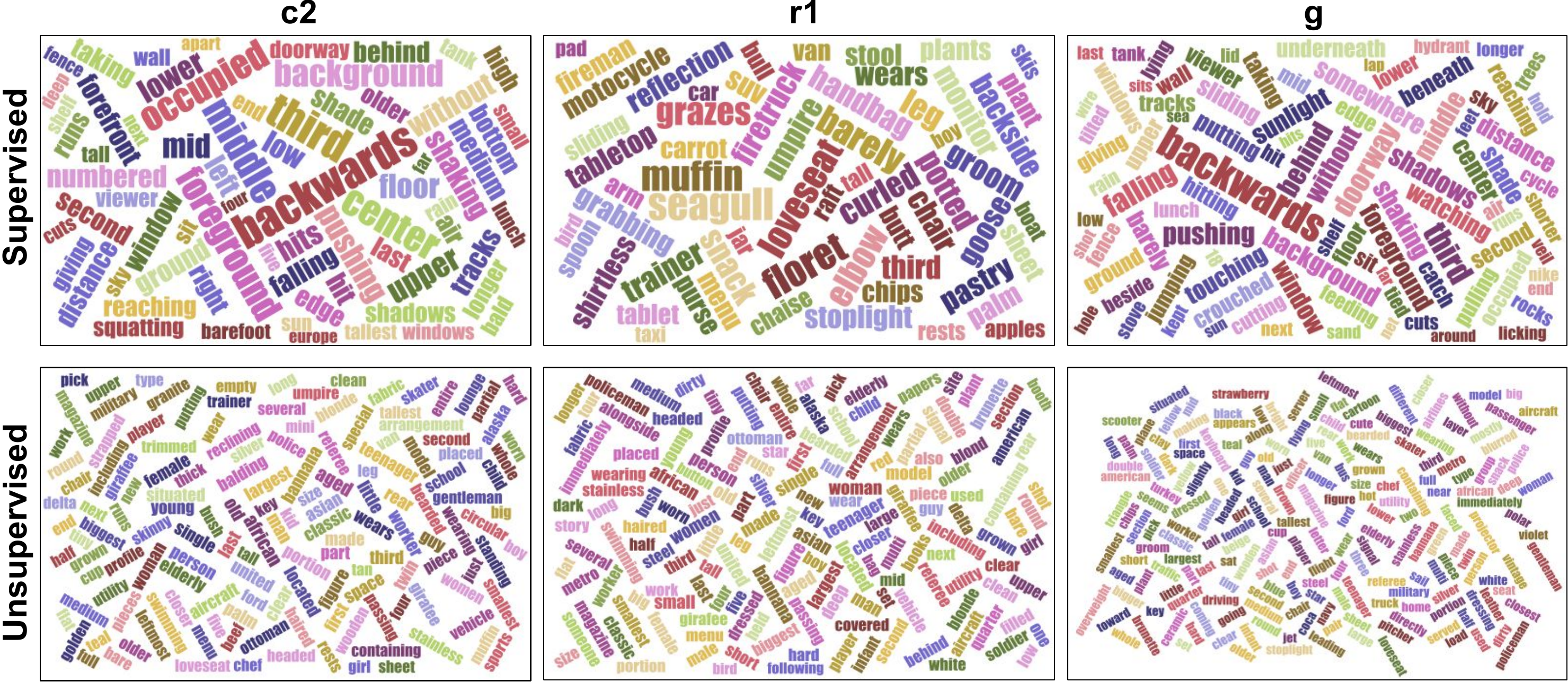}
	\caption{Word cloud visualizations of cue-specific word attention $\alpha$ in Eq.~\ref{eq:10} of context-cue (c2), referent-cue (r1), and generic-cue (g) using supervised (top row) and unsupervised training (bottom row) on RefCOCOg. Without supervision, it is difficult to discover meaningful language compositions.}
\vspace{-3mm}
\label{fig:6}
\end{figure}

\textbf{Unsupervised Relation Discovery}. Although we demonstrated that VC improves the unsupervised grounding by modeling context, we believe that there is still a large space for improving the quality of modeling the context. As the failure examples shown in Figure~\eqref{fig:7}, 1) many context estimations are still out of the scope of the expression, \eg, we may localize the ``cup'' and ``table'' as context even though the expression is ``woman with green t-shirt''; 2) we may mistake due to the wrong comprehension of the relations, \eg, ``right'' as ``left'', even if the objects belong to the same category, \eg, ``elephant''. For further investigation, Figure~\ref{fig:6} visualizes the cue-specific word attentions in supervised and unsupervised settings. The almost identical word attentions in unsupervised setting reflect the fact that the relation modeling between referent and context is not as successful as in supervised setting. This inspires us to exploit stronger prior knowledge such as language structure~\cite{xiao2017cvpr} and spatial configurations~\cite{zhang2017ppr,wei2017object}.

\section{Conclusions}
We focused on the task of grounding referring expressions in images and discussed that the key problem is how to model the complex context, which is not effectively resolved by the multiple instance learning framework used in prior works. Towards this challenge, we introduced the Variational Context model, where the variational lower-bound can be interpreted by the reciprocity between the referent and context: given any of which can help to localize the other, and hence is expected to significantly reduce the context complexity in a principled way. We implemented the model using cue-specific language-vision embedding network that can be efficiently trained end-to-end. We validated the effectiveness of this reciprocity by promising supervised and unsupervised experiments on four benchmarks. Moving forward, we are going to 1) incorporate expression language generation in the variational framework, 2) use more structural features of language rather than word attentions, and 3) further investigate the potential of our model in the unsupervised referring expression grounding.

\section{Supplementary Material}

\setcounter{equation}{13}
\subsection{Derivation of Eq. (3)}
Recall that we are going to transform the log-sum objective function in Eq. (2) to sum-log for tractable training. Without loss of generality, we omit the conditional $L$ in the derivation. By using the concavity of the log function:
\begin{equation}
\log((1-\alpha) x + \alpha y) \geq (1-\alpha) \log(x) + \alpha \log(y),
\end{equation}
to the following rewriting of Eq. (2): 
\begin{equation}
\begin{split}
&\log \sum\limits_z p(x, z) = log\left(\sum\limits_z \frac{q(z|x)p(x,z)}{q(z|x)}\right)\\
&\geq\sum\limits_z q(z|x) \log \frac{p(x,z)}{q(z|x)}\\
&=\sum\limits_z q(z|x)\log p(x|z)p(z) - \sum\limits_z q(z|x) \log q(z|x)\\
&=\sum\limits_z q(z|x)\log p(x|z) +\sum\limits_z q(z|x)\log p(z)\\
&~~~~~~~~~~~~~~~~~~~~~~~~~~~~~~~~~~~~~~~~~~~~~~~~~-\sum\limits_z q(z|x)\log q(z|x)\\
&= \sum\limits_z q(z|x)\log p(x|z) -\sum\limits_z q(z|x) \frac{q(z|x)}{p(z)}\\
&= \mathbb{E}_{z\sim q(z|x)}\log p(x|z) - KL\left(q(z|x)||p(z)\right) = \mathcal{Q}(x)
\end{split}
\end{equation}

\subsection{More Examples on Language Features}
Figure~\ref{fig:9} shows more cue-specific language features on RefCOCOg. 

\subsection{More Results on Unsupervised Grounding}
\subsection{External Parsers}
\setcounter{figure}{7}
\begin{figure}
	\centering
	\includegraphics[width=.7\linewidth]{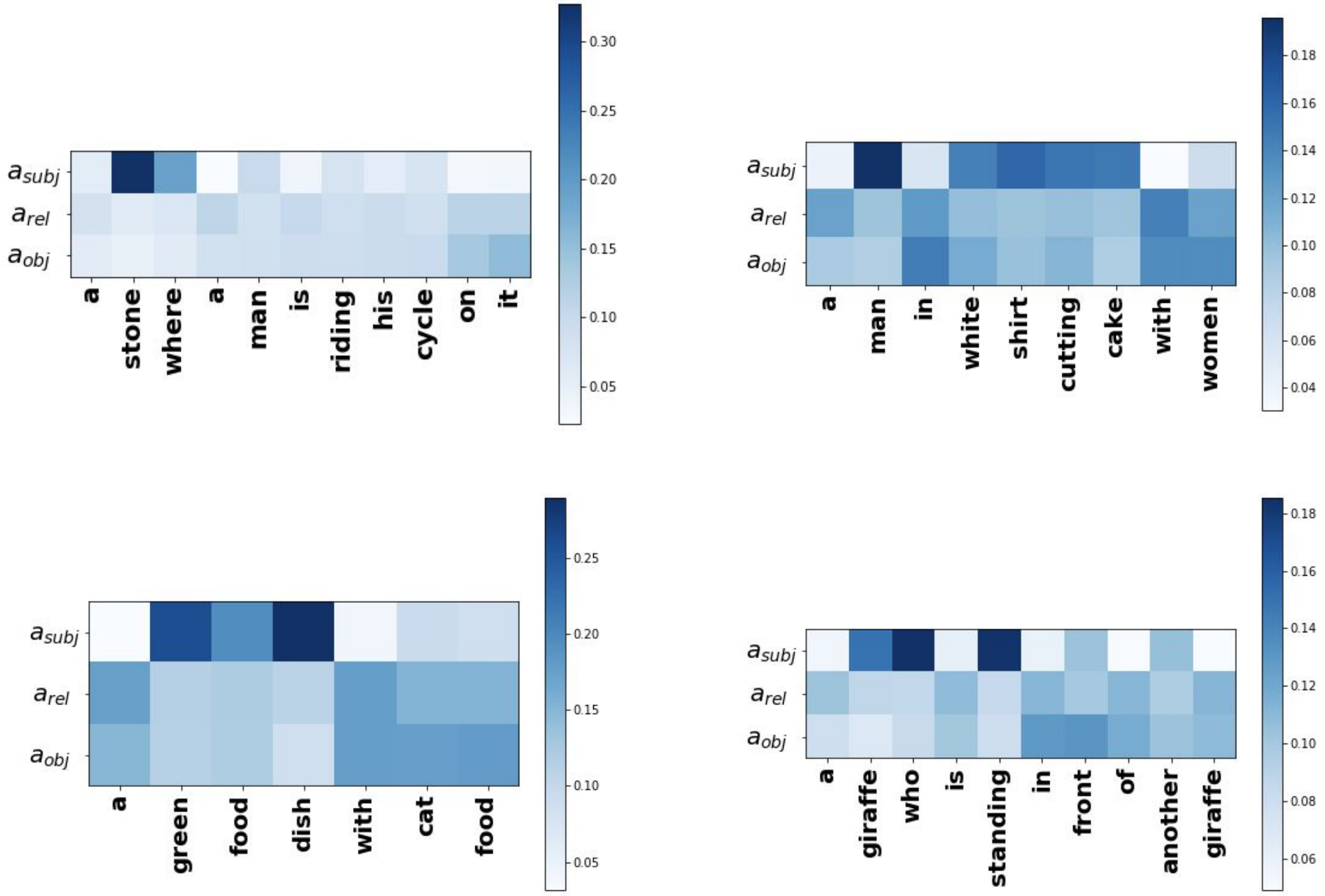}
	\caption{Some parsing examples from [11].}
\label{fig:8}
\end{figure}
As discussed in Section 5.4 that the language feature in unsupervised VC is not as good as that in the supervised setting. An alternative is to use external NLP parsers to obtain the compositions. However, conventional parsers (\eg, Standford Dependency) are observed to be suboptimal to the visual grounding task [11]. Therefore, we adopt the parser jointly trained on the referring expression grounding task [11]. As illustrated in Figure~\ref{fig:8}, this parser assigns word-level attention weights of subject, relation, and object. In particular, we consider the language features of c1, r2 as the average word embeddings, c2 as the relation weights, r1 as the subject weights, g as the object weights. Table~\ref{tab:4} shows the performances on unsupervised grounding. We can see that there is no significant improvement of VC w/ parser over VC w/o $\alpha$. 
\subsection{More Qualitative Results}
\begin{figure*}
	\centering
	\includegraphics[width=1\linewidth]{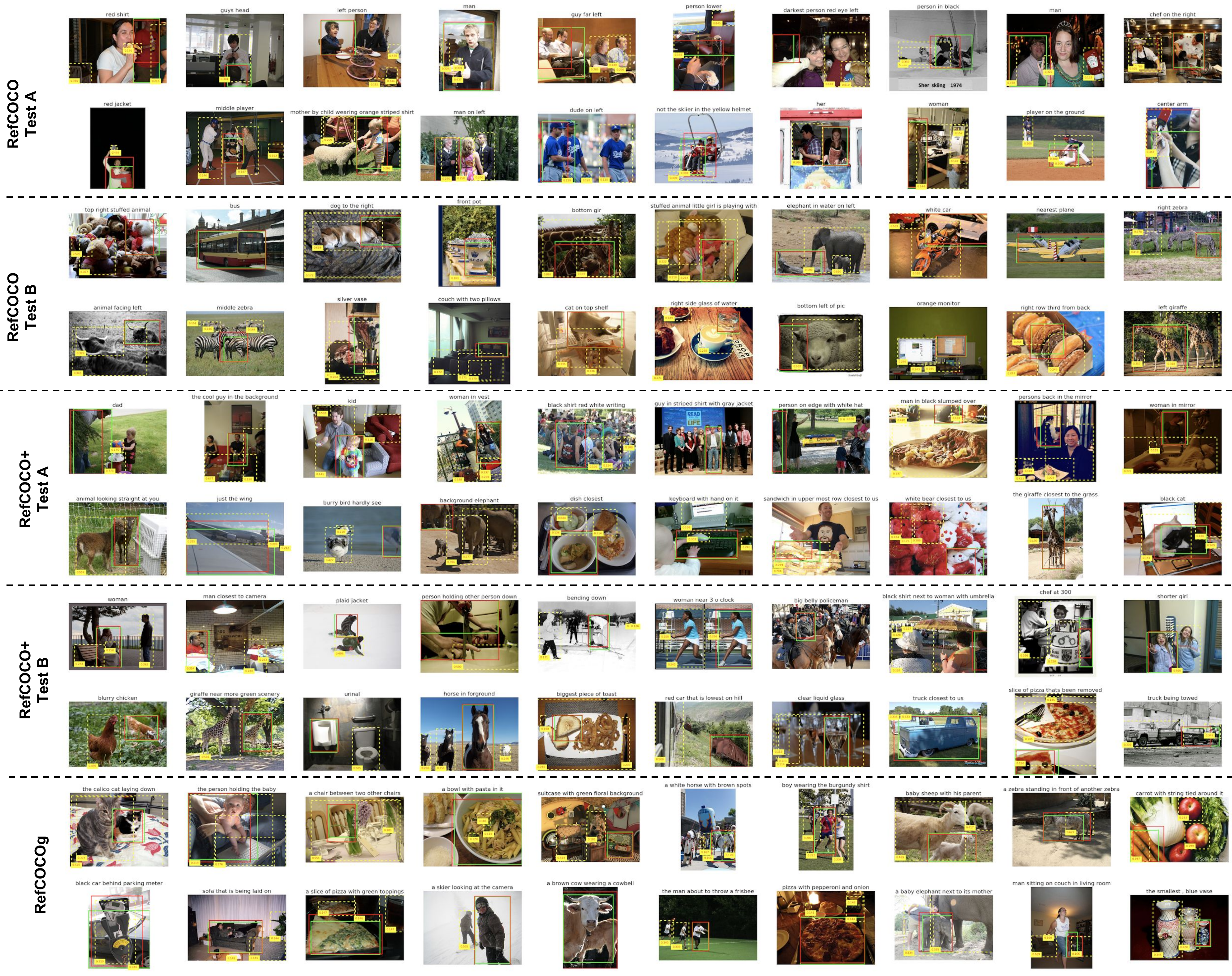}
	\caption{Qualitative results on RefCOCO, RefCOCO, RefCOCO+, and RefCOCOg. All of them are from det bounding boxes. The first line denotes the supervised results and the second line denotes the unsupervised results. Solid red: referent grounding; solid green: ground truth; dashed yellow: context grounding. We only display top 3 context objects with the context grounding probability $>0.1$. Please zoom in.}
\label{fig:10}
\end{figure*}

Figure~\ref{fig:10} shows more qualitative results on supervised and unsupervised grounding results on RefCOCO, RefCOCO+, and RefCOCOg.

\setcounter{table}{3}
\begin{table*}[t]
\centering
\caption{Unsupervised grounding performances (\%) of VC w/ or w/o parsers on the four datasets.}
\label{tab:4}
\begin{tabular}{|c|c|c|c|c|c|c|}
\hline
GT           & RefCLEF & RefCOCO TestA & RefCOCO TestB & RefCOCO+TestA & RefCOCO+TestB & RefCOCOg \\ \hline
VC w/ parser & 22.57 & 23.00 & 27.50 & 24.69 & 28.96 & 30.73    \\ \hline
VC           & 21.06 & 17.34 & 20.98 & 23.24 & 24.91 & 33.79 \\ \hline
VC w/o $\alpha$     & 20.72    & 33.29 & 30.13 & 34.60 & 31.58 & 30.26 \\ \hline
DET          & RefCLEF & RefCOCO TestA & RefCOCO TestB & RefCOCO+ TestA & RefCOCO+ TestB & RefCOCOg \\ \hline\hline
VC w/ parser & 14.90    & 24.07 & 25.12 & 26.18 & 26.92 & 30.64    \\ \hline
VC           & 14.11 & 20.91 & 21.77 & 25.79 & 25.54 & 33.66 \\ \hline
VC w/o $\alpha$     & 14.50 & 32.68 & 27.22 & 34.68 & 28.10 & 29.65 \\ \hline
\end{tabular}
\end{table*}

 \begin{figure*}
	\centering
	\includegraphics[width=1\linewidth]{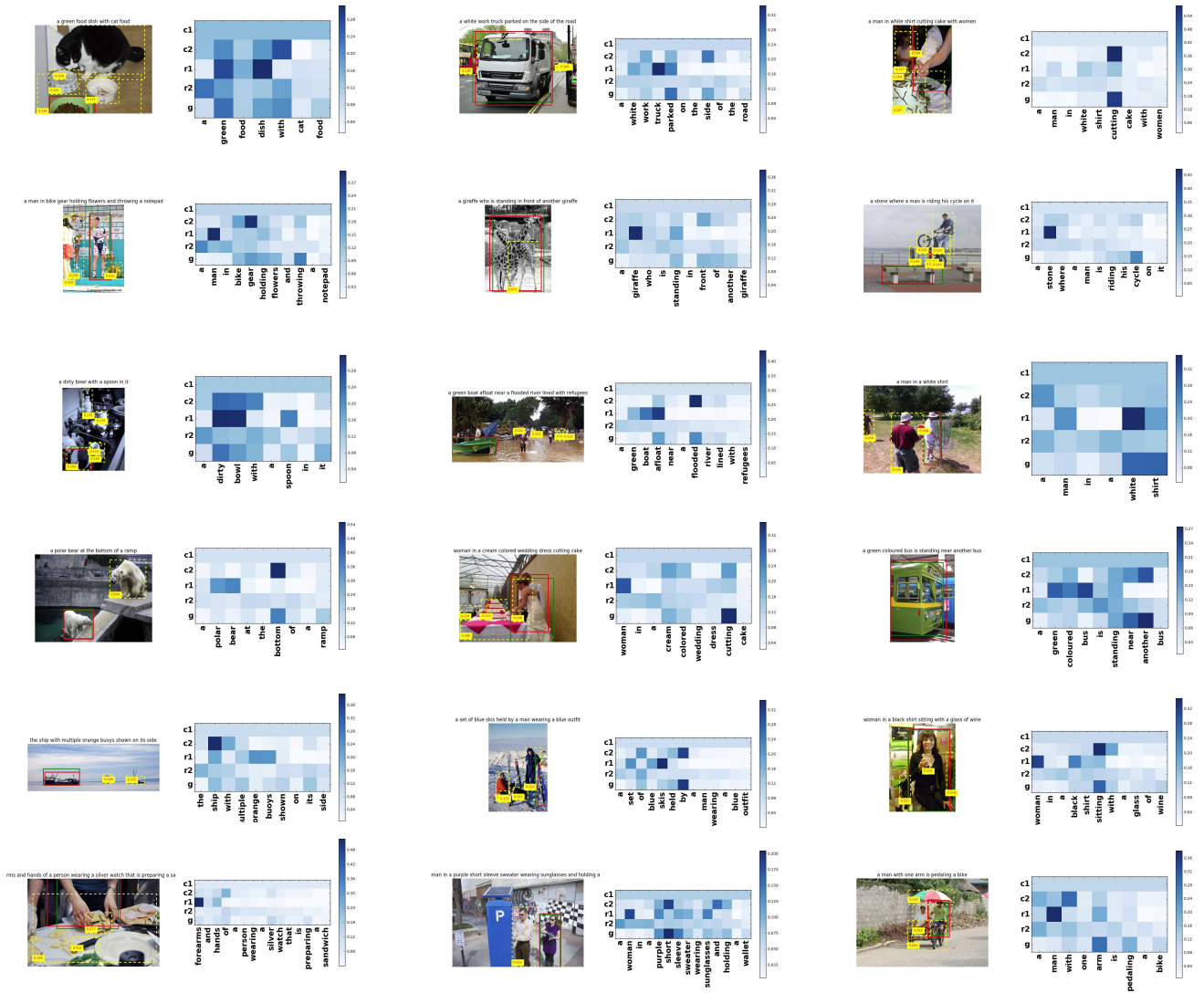}
	\caption{Qualitative results of the cue-specific language features on RefCOCOg.}
\label{fig:9}
\end{figure*}

{\footnotesize
\bibliographystyle{ieee}
\bibliography{egbib}

\begin{thebibliography}{10}\itemsep=-1pt

\bibitem{antol2015vqa}
S.~Antol, A.~Agrawal, J.~Lu, M.~Mitchell, D.~Batra, C.~Lawrence~Zitnick, and
  D.~Parikh.
\newblock Vqa: Visual question answering.
\newblock In {\em ICCV}, 2015.

\bibitem{ba2014multiple}
J.~Ba, V.~Mnih, and K.~Kavukcuoglu.
\newblock Multiple object recognition with visual attention.
\newblock In {\em ICLR}, 2015.

\bibitem{bahdanau2014neural}
D.~Bahdanau, K.~Cho, and Y.~Bengio.
\newblock Neural machine translation by jointly learning to align and
  translate.
\newblock 2015.

\bibitem{cho2014properties}
K.~Cho, B.~Van~Merri{\"e}nboer, D.~Bahdanau, and Y.~Bengio.
\newblock On the properties of neural machine translation: Encoder-decoder
  approaches.
\newblock {\em arXiv preprint arXiv:1409.1259}, 2014.

\bibitem{Dai_2017_CVPR}
B.~Dai, Y.~Zhang, and D.~Lin.
\newblock Detecting visual relationships with deep relational networks.
\newblock In {\em CVPR}, 2017.

\bibitem{visdial_rl}
A.~Das, S.~Kottur, J.~M. Moura, S.~Lee, and D.~Batra.
\newblock Learning cooperative visual dialog agents with deep reinforcement
  learning.
\newblock In {\em ICCV}, 2017.

\bibitem{dietterich1997solving}
T.~G. Dietterich, R.~H. Lathrop, and T.~Lozano-P{\'e}rez.
\newblock Solving the multiple instance problem with axis-parallel rectangles.
\newblock {\em Artificial intelligence}, 1997.

\bibitem{fox2012tutorial}
C.~W. Fox and S.~J. Roberts.
\newblock A tutorial on variational bayesian inference.
\newblock {\em Artificial intelligence review}, 2012.

\bibitem{glorot2010understanding}
X.~Glorot and Y.~Bengio.
\newblock Understanding the difficulty of training deep feedforward neural
  networks.
\newblock In {\em ICAIS}, 2010.

\bibitem{golland2010game}
D.~Golland, P.~Liang, and D.~Klein.
\newblock A game-theoretic approach to generating spatial descriptions.
\newblock In {\em EMNLP}, 2010.

\bibitem{hu2016modeling}
R.~Hu, M.~Rohrbach, J.~Andreas, T.~Darrell, and K.~Saenko.
\newblock Modeling relationships in referential expressions with compositional
  modular networks.
\newblock In {\em CVPR}, 2017.

\bibitem{hu2016natural}
R.~Hu, H.~Xu, M.~Rohrbach, J.~Feng, K.~Saenko, and T.~Darrell.
\newblock Natural language object retrieval.
\newblock In {\em CVPR}, 2016.

\bibitem{ioffe2015batch}
S.~Ioffe and C.~Szegedy.
\newblock Batch normalization: Accelerating deep network training by reducing
  internal covariate shift.
\newblock In {\em ICML}, 2015.

\bibitem{kazemzadeh2014referitgame}
S.~Kazemzadeh, V.~Ordonez, M.~Matten, and T.~L. Berg.
\newblock Referitgame: Referring to objects in photographs of natural scenes.
\newblock In {\em EMNLP}, 2014.

\bibitem{kingma2013auto}
D.~P. Kingma and M.~Welling.
\newblock Auto-encoding variational bayes.
\newblock In {\em ICLR}, 2014.

\bibitem{li2017visual}
Y.~Li, N.~Duan, B.~Zhou, X.~Chu, W.~Ouyang, X.~Wang, and M.~Zhou.
\newblock Visual question generation as dual task of visual question answering.
\newblock In {\em CVPR}, 2018.

\bibitem{li2017vip}
Y.~Li, W.~Ouyang, and X.~Wang.
\newblock Vip-cnn: Visual phrase guided convolutional neural network.
\newblock In {\em CVPR}, 2017.

\bibitem{lin2014microsoft}
T.-Y. Lin, M.~Maire, S.~Belongie, J.~Hays, P.~Perona, D.~Ramanan,
  P.~Doll{\'a}r, and C.~L. Zitnick.
\newblock Microsoft coco: Common objects in context.
\newblock In {\em ECCV}, 2014.

\bibitem{liureferring}
J.~Liu, L.~Wang, and M.-H. Yang.
\newblock Referring expression generation and comprehension via attributes.
\newblock In {\em ICCV}, 2017.

\bibitem{liu2016ssd}
W.~Liu, D.~Anguelov, D.~Erhan, C.~Szegedy, S.~Reed, C.-Y. Fu, and A.~C. Berg.
\newblock Ssd: Single shot multibox detector.
\newblock In {\em ECCV}, 2016.

\bibitem{lu2016visual}
C.~Lu, R.~Krishna, M.~Bernstein, and L.~Fei-Fei.
\newblock Visual relationship detection with language priors.
\newblock In {\em ECCV}, 2016.

\bibitem{lu2016hierarchical}
J.~Lu, J.~Yang, D.~Batra, and D.~Parikh.
\newblock Hierarchical question-image co-attention for visual question
  answering.
\newblock In {\em NIPS}, 2016.

\bibitem{luo2017comprehension}
R.~Luo and G.~Shakhnarovich.
\newblock Comprehension-guided referring expressions.
\newblock In {\em CVPR}, 2017.

\bibitem{aae}
A.~Makhzani, J.~Shlens, N.~Jaitly, and I.~J. Goodfellow.
\newblock Adversarial autoencoders.
\newblock In {\em ICLR Workshop}, 2016.

\bibitem{mao2016generation}
J.~Mao, J.~Huang, A.~Toshev, O.~Camburu, A.~L. Yuille, and K.~Murphy.
\newblock Generation and comprehension of unambiguous object descriptions.
\newblock In {\em CVPR}, 2016.

\bibitem{nagaraja2016modeling}
V.~K. Nagaraja, V.~I. Morariu, and L.~S. Davis.
\newblock Modeling context between objects for referring expression
  understanding.
\newblock In {\em ECCV}, 2016.

\bibitem{pennington2014glove}
J.~Pennington, R.~Socher, and C.~Manning.
\newblock Glove: Global vectors for word representation.
\newblock In {\em EMNLP}, 2014.

\bibitem{plummer2016phrase}
B.~A. Plummer, A.~Mallya, C.~M. Cervantes, J.~Hockenmaier, and S.~Lazebnik.
\newblock Phrase localization and visual relationship detection with
  comprehensive linguistic cues.
\newblock In {\em ICCV}, 2017.

\bibitem{plummer2015flickr30k}
B.~A. Plummer, L.~Wang, C.~M. Cervantes, J.~C. Caicedo, J.~Hockenmaier, and
  S.~Lazebnik.
\newblock Flickr30k entities: Collecting region-to-phrase correspondences for
  richer image-to-sentence models.
\newblock In {\em ICCV}, 2015.

\bibitem{redmon2016yolo9000}
J.~Redmon and A.~Farhadi.
\newblock Yolo9000: better, faster, stronger.
\newblock In {\em CVPR}, 2017.

\bibitem{ren2015faster}
S.~Ren, K.~He, R.~Girshick, and J.~Sun.
\newblock Faster r-cnn: Towards real-time object detection with region proposal
  networks.
\newblock In {\em NIPS}, 2015.

\bibitem{rohrbach2016grounding}
A.~Rohrbach, M.~Rohrbach, R.~Hu, T.~Darrell, and B.~Schiele.
\newblock Grounding of textual phrases in images by reconstruction.
\newblock In {\em ECCV}, 2016.

\bibitem{schuster1997bidirectional}
M.~Schuster and K.~K. Paliwal.
\newblock Bidirectional recurrent neural networks.
\newblock {\em TSP}, 1997.

\bibitem{schuster2015generating}
S.~Schuster, R.~Krishna, A.~Chang, L.~Fei-Fei, and C.~D. Manning.
\newblock Generating semantically precise scene graphs from textual
  descriptions for improved image retrieval.
\newblock In {\em Workshop on Vision and Language}, 2015.

\bibitem{sohn2015learning}
K.~Sohn, H.~Lee, and X.~Yan.
\newblock Learning structured output representation using deep conditional
  generative models.
\newblock In {\em NIPS}, 2015.

\bibitem{sun2017domain}
Q.~Sun, B.~Schiele, and M.~Fritz.
\newblock A domain based approach to social relation recognition.
\newblock In {\em CVPR}, 2017.

\bibitem{thomas2014meaning}
J.~A. Thomas.
\newblock {\em Meaning in interaction: An introduction to pragmatics}.
\newblock Routledge, 2014.

\bibitem{thomason2017guiding}
J.~Thomason, J.~Sinapov, and R.~Mooney.
\newblock Guiding interaction behaviors for multi-modal grounded language
  learning.
\newblock In {\em Proceedings of the First Workshop on Language Grounding for
  Robotics}, pages 20--24, 2017.

\bibitem{wei2017object}
Y.~Wei, J.~Feng, X.~Liang, C.~Ming-Ming, Y.~Zhao, and S.~Yan.
\newblock Object region mining with adversarial erasing: A simple
  classification to semantic segmentation approach.
\newblock In {\em CVPR}, 2017.

\bibitem{williams1992simple}
R.~J. Williams.
\newblock Simple statistical gradient-following algorithms for connectionist
  reinforcement learning.
\newblock In {\em Machine Learning}. 1992.

\bibitem{xiao2017cvpr}
F.~Xiao, L.~Sigal, and Y.-J. Lee.
\newblock Weakly-supervised visual grounding of phrases with linguistic
  structures.
\newblock In {\em CVPR}, 2017.

\bibitem{xu2015show}
K.~Xu, J.~Ba, R.~Kiros, K.~Cho, A.~Courville, R.~Salakhudinov, R.~Zemel, and
  Y.~Bengio.
\newblock Show, attend and tell: Neural image caption generation with visual
  attention.
\newblock In {\em ICML}, 2015.

\bibitem{xue2016visual}
T.~Xue, J.~Wu, K.~Bouman, and B.~Freeman.
\newblock Visual dynamics: Probabilistic future frame synthesis via cross
  convolutional networks.
\newblock In {\em NIPS}, 2016.

\bibitem{yan2016attribute2image}
X.~Yan, J.~Yang, K.~Sohn, and H.~Lee.
\newblock Attribute2image: Conditional image generation from visual attributes.
\newblock In {\em ECCV}, 2016.

\bibitem{yu2016modeling}
L.~Yu, P.~Poirson, S.~Yang, A.~C. Berg, and T.~L. Berg.
\newblock Modeling context in referring expressions.
\newblock In {\em ECCV}, 2016.

\bibitem{yu2016joint}
L.~Yu, H.~Tan, M.~Bansal, and T.~L. Berg.
\newblock A joint speaker-listener-reinforcer model for referring expressions.
\newblock In {\em ICCV}, 2017.

\bibitem{zhang2016vtranse}
H.~Zhang, Z.~Kyaw, S.-F. Chang, and T.-S. Chua.
\newblock Visual translation embedding network for visual relation detection.
\newblock In {\em CVPR}, 2017.

\bibitem{zhang2017ppr}
H.~Zhang, Z.~Kyaw, J.~Yu, and S.-F. Chang.
\newblock Ppr-fcn: Weakly supervised visual relation detection via parallel
  pairwise r-fcn.
\newblock In {\em ICCV}, 2017.

\bibitem{zhao2017video}
Z.~Zhao, Q.~Yang, D.~Cai, X.~He, and Y.~Zhuang.
\newblock Video question answering via hierarchical spatio-temporal attention
  networks.
\newblock In {\em International Joint Conference on Artificial Intelligence
  (IJCAI)}, volume~2, 2017.

\bibitem{zitnick2014edge}
C.~L. Zitnick and P.~Doll{\'a}r.
\newblock Edge boxes: Locating object proposals from edges.
\newblock In {\em ECCV}, 2014.

\end{thebibliography}
}

\end{document}